# Soft Pneumatic Actuators for Soft Robotics: A Motion-Based Review of Actuation Mechanisms and Performance Trade-offs

Mohammed Abboodi; https://orcid.org/0009-0002-4645-0026

**Abstract:** Soft pneumatic actuators are widely used in soft robotics because they can produce large motions while remaining compliant enough to interact safely with objects, environments, and the human body. However, their performance is not solely determined by pressure. Instead, the response depends on the way the actuator is built, including the shape of its chambers, the placement of reinforcements, the use of folds, material stiffness, and the constraints that guide its deformation. As the literature has expanded, it has become more difficult to determine which mechanism is most suitable for a given application and which reported results can be compared across studies. This review examines soft pneumatic actuators according to the design strategies used to generate four motion classes: linear, bending, twisting, and omnidirectional actuation. For each class, it analyzes the structural features that define the deformation path, including braid angle, fold geometry, fiber orientation, chamber arrangement, structural asymmetry, and internal constraint layers. It then discusses how the design choice affect motion output, force generation, air demand, repeatability, durability, fabrication difficulty, and robotic integration. The review further identifies key conditions that must be considered when selecting or comparing actuators, including pressure, loading condition, actuator size, pneumatic supply, and hysteresis This approach helps explain why actuators with similar motion outputs may differ substantially in design requirements, pneumatic demand, and practical suitability. It also highlights the design priorities needed for compact, efficient, repeatable, and deployable soft pneumatic systems in wearable, biomedical, and mobile robotic applications.

## 1. INTRODUCTION

Soft robotics has become an important area of robotics because compliant structures can deform continuously, adapt to irregular surfaces, and distribute contact forces during interaction. These properties are especially useful for handling delicate objects, operating in uncertain environments, and interacting safely with the human body, where compliance is a basic design requirement rather than an added feature [1]. In these systems, the actuator is central to performance because it not only produces motion but also influences the load path, deformation mode, stiffness distribution, and contact behavior of the robot.

Most soft robotic motions can be described in terms of a few basic modes: linear, bending, twisting, and omnidirectional motion. More complex behaviors usually come from combining these modes in serial, parallel, or hybrid configurations. In soft pneumatic actuators, the main challenge is not simply to inflate a chamber, but to guide the deformation into a useful and repeatable path. Without proper constraints, pressure can produce radial ballooning, unintended bending, parasitic twisting, or instability under load. For this reason, chamber geometry, reinforcement layout, folds, structural asymmetry, and internal constraints are central design features that determine how pneumatic input is converted into useful motion.

Several soft actuation technologies have been developed, but each carries practical limitations. Shape memory alloy actuators rely on thermally induced phase transformation, which can lead to slow cooling response, hysteresis, and temperature-dependent behavior [2]. Twisted and coiled polymer actuators also depend on thermal input, so their use is affected by cooling requirements and electrothermal efficiency losses [3]. Dielectric elastomer actuators can provide high strain and fast response, but they require high driving voltages and may suffer from electrical breakdown or material degradation during repeated use [4]. Cable-driven soft systems offer compact transmission and direct motor control, yet friction, backlash, hysteresis, and slippage at soft interfaces can reduce force transmission and tracking accuracy [5], [6], [7]. Soft pneumatic actuation remains widely used because it combines large deformation, compliance, and relatively simple construction. Its main challenge is usually not producing motion, but supplying and regulating air through pumps, valves, regulators, tubing, and seals, which can strongly affect bandwidth, portability, and measured performance [8].

The growing number of soft pneumatic actuator designs has made mechanism selection more difficult. Similar motions can be produced by very different architectures, including pneumatic muscles, bellows, PneuNets, fiber-reinforced chambers, textile actuators, and sleeve-based mechanisms. These designs may appear comparable when judged only by stroke, bending angle, twisting range, or workspace, but their performance can change substantially with actuator size, pressure range, loading condition, flow capacity, fabrication method, and test setup. For this reason, motion output alone is not enough to evaluate actuator performance unless the pneumatic and experimental conditions are also considered.

Existing reviews have already examined the design, fabrication, modeling, sensing, control, and applications of soft pneumatic actuators [8], [9]. Others have examined specific families such as bellows, pneumatic artificial muscles, and fabric-based pneumatic actuators [10], [11] [12]. This review looks at how different actuator designs produce linear, bending, twisting, and omnidirectional motion, and what compromises come with each design choice. Instead of simply listing actuator types, this review explains how selected designs generate motion and why some mechanisms are better suited than others for particular soft robotic applications.

This review is organized around three questions. How do soft pneumatic actuators convert pressure into linear, bending, twisting, or omnidirectional motion? What is gained or lost when actuator designs are judged by force, stroke, air demand, repeatability, durability, fabrication, and integration? The review also asks when results from different studies can be compared fairly, and when they are too dependent on pressure range, loading condition, pneumatic supply, or test setup. These questions keep the review focused on design decisions and on the requirements for compact, efficient, and practical soft pneumatic systems.

# 2. Linear actuators

## 2.1 Pneumatic artificial muscles

Pneumatic artificial muscles (PAMs) are among the most established soft pneumatic actuators for producing linear motion. In these actuators, pressure first expands a compliant chamber, while a braided, pleated, netted, or embedded constraint redirects that expansion into axial shortening and tensile force [13]. The McKibben muscle shows this mechanism clearly: an elastomeric bladder inflates inside a helical braided sleeve, the sleeve angle changes as the diameter increases, and the muscle contracts along its length [13], [14]. This gives PAMs a useful combination of low mass and high tensile output. The same coupling, however, makes their response difficult to predict. The pressure–force relation changes with braid geometry, interfacial friction, viscoelasticity, and end-region deformation, so the measured behavior is nonlinear and often history dependent [15].

Among PAM designs, the braided McKibben muscle is the most widely used and most frequently studied. In ideal models, its contraction and force output are mainly determined by the initial braid angle, actuator length, and radius [16]. Actual behavior is more complicated because the bladder and sleeve do not move as perfectly coupled parts. Sliding at the bladder–sleeve interface, friction between yarns, membrane stiffness, sleeve pretension, and end termination can all affect the measured response [13], [15] ,[16]. This is why braided PAMs can generate strong linear motion but remain difficult to model and compare across studies. Their performance depends not only on pressure and geometry, but also on assembly details and material interactions that are often not fully reported.

Pleated PAMs address part of this problem by changing the deformation mechanism. Instead of relying mainly on sliding between a bladder and a braided sleeve, the actuator contracts through folding and unfolding of a pleated membrane (Figure 1a) [17], [18]. This reduces material stretching and dry friction, which can improve repeatability compared with conventional braided muscles. The reported capability of this approach is notable: Daerden and co-workers described a 60 g pleated muscle capable of producing up to 3500 N with a contraction ratio of 42% [19]. The trade-off is that the design becomes more dependent on the quality of the pleat geometry. Pleat depth, fold radius, membrane thickness, seam stiffness, and end reinforcement influence local strain and fatigue. Pleated PAMs therefore reduce some interface losses, but they shift the durability problem toward the folds themselves.

Netted PAMs replace the dense braided sleeve with a more open mesh or harness (Figure 1b). This changes how the constraint interacts with the membrane. In some designs, such as the Kukolj-type muscle, the mesh does not fully contact the membrane at low load; the gap closes only when the external load becomes large enough [11]. This behavior can be useful when the actuator needs to start with a soft response and then become stiffer under load. However, it can also create a dead zone and reduce precision at low loads. The response therefore depends strongly on mesh pretension, aperture size, aspect ratio, and operating load. For netted muscles, peak force and contraction ratio are not enough to describe performance; the way the mesh engages the membrane is also part of the actuator behavior.

Embedded PAMs integrate the reinforcement into the actuator wall instead of using a separate outer sleeve. This design can reduce bladder-sleeve sliding and improve sealing, since the constraint that carries the load is built into the actuator structure [20]. The Festo Fluidic Muscle is one example; for a standard size, it is specified to operate up to 0.6 MPa and produce a theoretical force of 1500 N [21], [22]. The advantage of this approach is more standardized and repeatable performance. The limitation is that the actuator cannot be easily adjusted after fabrication. Once the reinforcement is embedded, the response depends on fiber alignment, bonding quality, defect control, and the way the end regions are designed.

Overall, PAMs show how strongly linear pneumatic actuation depends on the way radial expansion is constrained. Braided, pleated, netted, and embedded designs all convert pressure into axial work, but they do so through different mechanical pathways. The comparison also shows why reported PAM performance should be interpreted with care. Maximum force or contraction ratio is not enough; pressure range, load condition, contraction level, hysteresis, and construction details are needed to understand whether two PAMs are genuinely comparable.

### 2.2 Bellows actuators

Bellows-based linear soft pneumatic actuators address a basic problem in symmetric inflatable chambers. When a plain chamber is pressurized, part of the input is often lost to radial expansion, or ballooning, instead of being directed into useful axial stroke [23], [24]. Bellows reduce this problem by adding geometry that constrains the wall. Corrugations, convolutions, or prescribed folds guide the deformation path, increase resistance to uncontrolled radial expansion, and make axial motion more repeatable [9], [10].

In conventional corrugated bellows, the operating principle is mainly geometric. The repeated convolutions allow the actuator to extend or contract along its axis while limiting radial growth compared with an unconstrained tube [23]. This gives bellows a practical advantage: they can generate linear motion without an external sleeve, braid, or linkage mechanism [8]. Their limitations, however, are not removed; they appear in a different form. Corrugated bellows can suffer from off-axis deformation, buckling under compression, and reduced stroke under load. Local strain also tends to concentrate near convolution roots and crests, which can shorten fatigue life when the fold curvature is too severe [25], [26].

The differences between U-shaped, V-shaped, toroidal, and flat-faced bellows are therefore not only changes in profile. Each geometry distributes curvature and stiffness differently [10]. Rounded U-shaped convolutions usually provide high axial compliance and relatively large stroke per pitch, which is useful when large extension is needed under modest loads [27]. The same compliance can also make the actuator more vulnerable to off-axis deformation and stroke loss when loaded [26], [28]. V-shaped convolutions create sharper hinge-like regions and can provide greater stroke within a compact actuator length, but they also increase strain concentration near the fold apex [29]. Toroidal bellows can reduce extreme curvature and improve lateral stability, although this usually increases axial stiffness and may require higher pressure, a larger effective area, or a longer active section to achieve comparable stroke [30], [31]. These examples show that bellows design is mainly a compromise between stroke, stiffness, stability, and fatigue resistance.

In conventional bellows, the most important design variables are effective area, wall thickness, convolution height, pitch, and number of convolutions [32], [33]These parameters control axial compliance, volume change per unit stroke, and stability under external load. In general, increasing the geometric constraint improves motion directionality and load-bearing capacity by limiting unwanted radial expansion, but it also increases stiffness and can reduce compliance. Bellows design is therefore not simply about maximizing displacement. It is about deciding where deformation should occur and how it should be stabilized.

Origami-derived bellows take this idea further by replacing smooth corrugations with defined fold patterns. In these actuators, the fold geometry prescribes the motion path more directly, so pressurization or vacuum can produce mainly axial extension or contraction. Early elastomeric origami actuators showed that accordion-like crease patterns could generate large axial motion with fast response, demonstrating that fold architecture can dominate the actuation behavior more effectively than uncontrolled membrane stretching [34]. Later fluid-driven origami muscles separated the function of the structure into an origami skeleton for kinematics, a compliant skin for sealing and pressure transfer, and the working fluid [35]. This makes the design problem more specific: the fold pattern must preserve axial motion under load while limiting parasitic twisting, crease damage, and mode switching.

Accordion-type patterns are often suitable when near-linear translation is required. Their repeated folds naturally bias deformation along the actuator axis and can maintain a nearly constant cross-section [34], [36]. Their response is also relatively interpretable because fold angle, pleat depth, and pitch can be adjusted to tune stroke and stiffness. In one vacuum-driven semi-soft actuator, accordion and Yoshimura patterns were used to achieve large contraction while preserving cross-sectional consistency. The actuator reached 40 mm contraction, 36% strain, and 124 N blocked force at −80 kPa, with stable displacement over 1000 cycles and no visible fracture in the core structure [36]. This result shows the value of fold-guided linear actuation, but it does not remove the durability concern. A comparison of 4D-printed accordion, Kresling, and Yoshimura tubes found that the accordion pattern produced the largest deformation, but cracks appeared at crease locations after only a few cycles in a brittle material system [37]. For fold-based actuators, large stroke is therefore not enough; the material and crease mechanics must also support repeated cycling.

Yoshimura and Miura-derived patterns offer different ways to guide axial deformation. Yoshimura-type folding uses a structured shell-buckling mode that can maintain a coherent tubular form and reduce uncontrolled ballooning. However, this benefit depends strongly on wall stiffness and boundary conditions. When the wall is not strong enough, Yoshimura tubes can buckle unfavorably, showing that guided buckling is useful only when competing instability modes are suppressed [37]. Miura-derived tubular patterns provide a more parameterized route to axial actuation and can simplify geometric modeling, but they may introduce shear or parasitic twist if the fold geometry is not carefully constrained [38]. This is a key point for linear actuators: a programmable fold pattern is only useful when the programmed motion matches the required degree of freedom.

Kresling-based bellows illustrate this trade-off more directly because their axial motion is inherently coupled with rotation. They can also switch between stable states through snapping [39]. This behavior can be useful in designs that require combined contraction and rotation, but it becomes a limitation when pure linear actuation is required unless rotation is mechanically constrained. Waterbomb-derived and related folded patterns allow contraction and force to be tuned through fold geometry, but comparison remains difficult when studies report only displacement. Blocked force, pressure range, hysteresis, repeatability, cycle life, and loading condition are needed to judge whether the reported performance comes from the actuator design itself or from the test conditions [35]

Bellows-based linear actuators therefore show that axial motion is not difficult to produce; the harder problem is controlling where the structure deforms and how stable that deformation remains under load. Conventional corrugated bellows use convolution geometry to suppress ballooning and stabilize axial stroke. Origami-derived bellows use fold patterns to prescribe the motion more explicitly. Both approaches can improve directionality, but both introduce trade-offs in stiffness, stability, fatigue, and manufacturability. For this reason, peak displacement alone is not a sufficient performance measure. Pressure range, external load, blocked force or load-dependent stroke, repeatability, hysteresis, and durability should be reported before bellows actuators from different studies are compared.

### 2.3 Fiber-reinforced soft actuators for linear motion (FREEs)

Fiber-reinforced soft actuators use anisotropic reinforcement to control how a pressurized elastomeric tube deforms. The fibers prevent the tube from expanding freely outward and guide the deformation according to the reinforcement layout [40]. For linear actuation, the design must direct radial inflation into axial extension or contraction while avoiding unwanted twisting or bending. This means that pressure alone does not define the motion; the fiber architecture determines how that pressure is converted into useful axial deformation.

The role of reinforcement becomes clear when comparing reinforced and unreinforced tubes. In an unreinforced tube, pressurization mainly produces radial expansion. In a fiber-reinforced actuator, the fibers are much stiffer along their own direction and therefore constrain the elastomeric wall. Connolly and co-workers modeled this behavior by treating the actuator as an elastomeric tube reinforced with fibers at an angle $\alpha$ relative to the actuator axis [40][41]. Changing $\alpha$ changes the coupling between radius and length, which determines whether the actuator extends, contracts, or develops coupled motion under pressure. When a primarily linear response is required, two symmetric fiber families, such as +α and −α, are commonly used to reduce parasitic torsion and promote axial motion [41]. Thus, linear motion in FREE-type actuators is not obtained simply by adding fibers; it depends on choosing a reinforcement layout that suppresses unwanted modes while preserving axial deformation.

An important development in this field was the move from experimental design toward model-based motion programming. Connolly et al. showed that fiber-reinforced soft actuators could be designed through a mechanics-based and optimization framework to match prescribed trajectories, including finger-motion replication [41], [42]. This changed the design problem from selecting a geometry empirically to solving an inverse problem in which the desired motion guides the actuator parameters. However, this programmability is not as straightforward as it may appear. Pressure–displacement and pressure–force responses remain nonlinear and state dependent, and they are affected by fiber engagement, local wall thickness, fiber–matrix interaction, viscoelasticity, and frictional losses. For this reason, simplified analytical models, continuum models, and data-driven models each offer a different balance between interpretability, calibration effort, and prediction accuracy[43]. A model calibrated for one actuator may not remain reliable when the geometry, fabrication method, or loading condition changes.

Manufacturing variability is the major limitation in fiber-reinforced actuators. Although fiber angle is often treated as the main design variable, the actual response also depends on winding accuracy, wall-thickness uniformity, and the quality of the fiber–elastomer bond. Even small deviations in these factors can change the pressure–stroke response, introduce coupled deformation, and reduce repeatability. This is why reinforcement density and geometry have been studied not only to program motion, but also to improve linearity and durability [44]. Fiber angle is therefore a useful design parameter, but it cannot guarantee predictable performance unless the reinforcement is placed consistently and remains bonded to the elastomer during repeated actuation.

Control also affects how these actuators perform in practice. Even with a suitable fiber layout, hysteresis, material nonlinearity, and actuator-to-actuator variation can reduce tracking accuracy [45]. Data-driven controllers can compensate for some of these effects, but they may be harder to interpret and less transferable when loading conditions or actuator properties change. The main strength of fiber-reinforced linear actuators is their ability to program motion through reinforcement geometry. Their main limitation is that this programmed behavior is sensitive to material properties, fabrication quality, and control assumptions.

Fiber-reinforced actuators show that linear pneumatic motion can be programmed effectively through reinforcement architecture, but they also show why geometry alone is not a complete design solution. Their performance depends on fiber angle, reinforcement symmetry, wall geometry, material response, fabrication consistency, and control strategy.

### 2.4 Linear soft sleeve actuators for axial motion (LSSAs)

Linear soft sleeve actuators (LSSA) represent a new approach for generating linear motion. In this actuator, linear motion does not come from the free expansion of a large chamber, but from the controlled folding and unfolding of a thin sleeve-like structure [23]. The key advantage is the small internal pneumatic volume. Because the inner and outer walls are close to each other and the actuator does not require a large central cavity, less air is needed during pressurization or vacuum actuation. This makes the design especially relevant for compact and wearable systems, where the size and capacity of the pneumatic supply are major constraints.

The motion of LSSAs comes from the way the folded sleeve wall is forced to deform. Positive pressure opens the folded wall and produces axial extension, while vacuum folds the structure back and produces axial contraction. In this sense, LSSAs share the same basic idea as bellows actuators: geometry is used to guide the motion; however, the sleeve form adds a specific problem. If the inner and outer walls are not tied together, the structure can deform radially instead of moving mainly along its axis. The inner wall may collapse inward, or the outer wall may expand outward, which distorts the cross-section and reduces useful axial output. Therefore, tie-restraining layers are essential because they connect the two walls, stabilize the cross-section, and help convert pneumatic input into axial motion [23]. For LSSAs, stable linear actuation depends not only on the folded geometry, but also on how well the internal restraints control the wall motion.

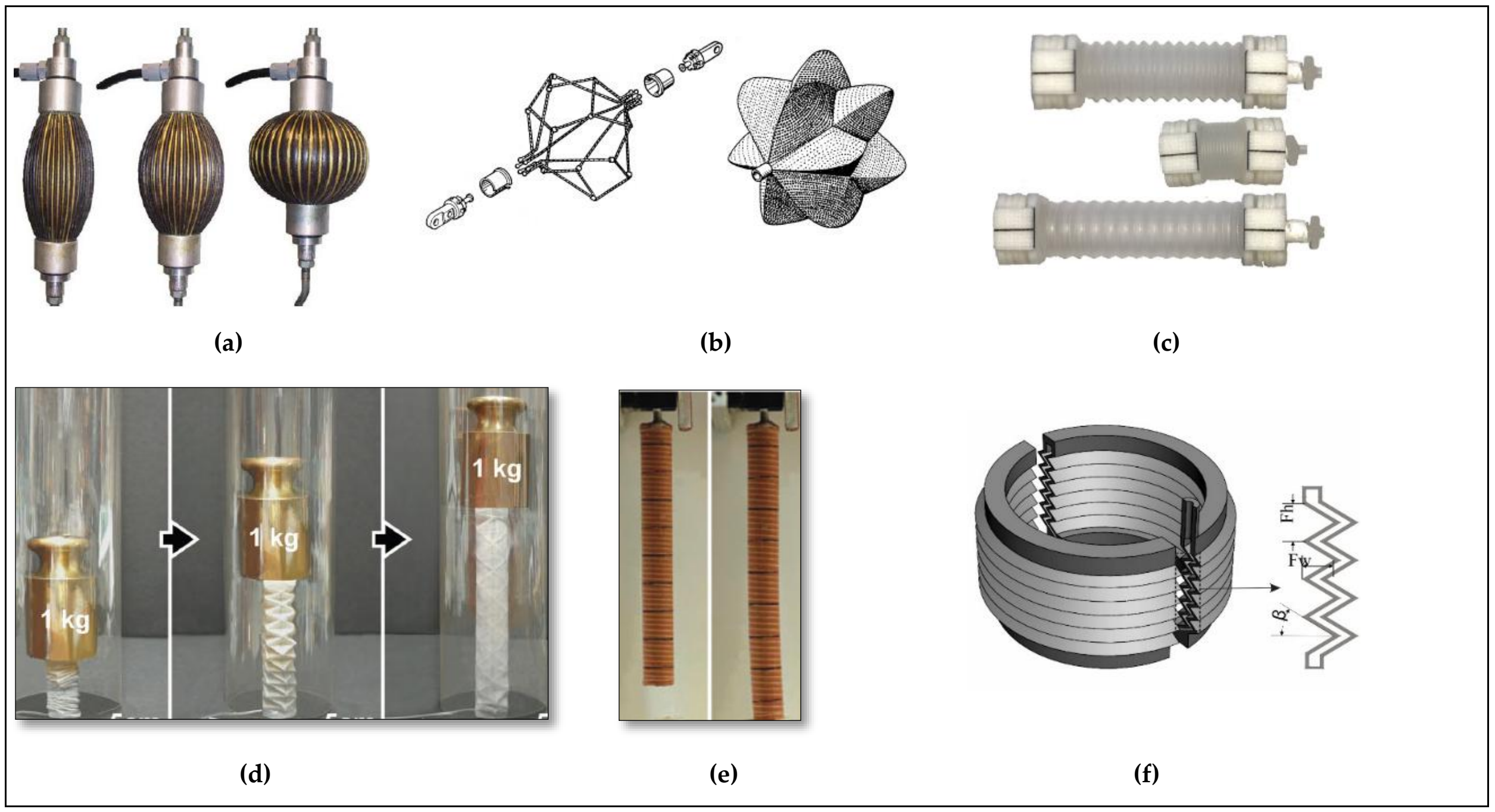

**Figure1:** Linear actuator: (a) Pleated PAM [46], (b) Netted muscles [47], (c) Linear bellows [48], (d) Origami linear actuator [34] , (e) Fiber-reinforced linear actuators [41], (f) LSSA [23]

The design progression reported in [23] shows this point clearly. The early ring-supported design improved linearity by using ring-like supports to limit cross-sectional distortion, but the added support also increased stiffness and reduced compliance. The lateral-support design replaced the rings with internal ties, improving flexibility while still controlling part of the radial deformation. The longitudinally supported folded design was the more effective step because it used longitudinal ties to stabilize the inner and outer walls along the actuator axis while allowing the folded structure to unfold under pressure [23]. The importance of this progression is not only the increase in stroke, but the improved balance between axial compliance and structural stability.

LSSAs offer three practical advantages over other linear soft pneumatic actuators. First of all, the sleeve form allows the actuation structure to be integrated around a limb-like or cylindrical body, which can reduce the need for external straps, braces, or textile transmission layers. Second, the same actuator can extend under positive pressure and contract under vacuum, giving bidirectional axial motion without requiring a separate antagonistic actuator. Third, the reduced pneumatic volume lowers airflow demand, which can improve response when the actuator is driven by a compact pneumatic supply [23]. These features make LSSAs particularly relevant for wearable and portable soft robotic systems.

However, these advantages depend strongly on the constraint design. The actuator response is sensitive to tie placement, tie stiffness, fold geometry, wall thickness, and manufacturing precision. Small changes in the internal restraints can change cross-sectional stability, alter the unfolding path, and reduce repeatability. Manufacturing is also a limiting factor because thin-walled airtight thermoplastic sleeves are difficult to produce consistently [23]. Therefore, the main trade-off in LSSAs is clear: reducing pneumatic volume and integrating bidirectional motion can improve compactness, but only if the internal constraint system prevents radial instability without making the actuator too stiff or too difficult to manufacture.

LSSAs show that sleeve-based pneumatic actuation is not simply a matter of adding folds to a thin chamber. Stable axial motion requires the correct level of internal constraint between the inner and outer walls. Too little constraint allows parasitic radial deformation; too much constraint reduces compliance and stroke. This makes LSSAs a useful example of a broader principle in soft pneumatic actuator design: the best geometry is not necessarily the most flexible one, but the one that provides enough constraint to preserve the intended motion

## 3. BENDING ACTUATORS

### 3.1 Pneumatic network actuators (PneuNets)

Pneumatic network actuators (PneuNets) are widely used for producing bending in soft robotics. They consist of a series of chambers embedded in a compliant elastomeric body. The operation of these actuators depends on patterned chambers embedded in a compliant elastomeric body; however, chamber count alone is not the main factor controlling performance. What matters is how chamber shape, wall thickness, and local stiffness force inflation to occur more on one side of the actuator than the other (Figure 2a). When pressure is applied, the thinner and more compliant chamber walls expand first. The unequal expansion elongation across the actuator body results strain mismatch which produces curvature [49] so bending motion is produced form of asymmetric expansion, not simple inflation of a soft chamber.

Early PneuNet studies showed that small design changes can strongly affect the bending response. Chamber geometry, material distribution, and chamber position relative to the free surfaces all influence where strain develops and how the actuator bends [49], [50], [51]. This is what makes PneuNets useful but also difficult to generalize. The design is simple, but the final motion depends on wall thickness, stiffness distribution, and boundary constraints. For this reason, finite-element analysis and experimental adjustment are still commonly needed in PneuNet design [49].

The fast PneuNet design shows why bending angle alone is a weak measure of performance (Figure 2b). Conventional PneuNets can reach large curvature, but often at the cost of a large chamber-volume change. In the reported comparison, full bending required nearly twenty times the initial chamber volume [51]. This demand reduces bandwidth, increases the required clearance, and raises material strain. The fast PneuNet addressed this limitation by adding gaps between adjacent chamber walls and making the internal walls more compliant than the outer walls. Under pressure, these internal walls expanded first and pushed neighboring chambers against each other, producing bending with little outward thickening [51]. The height change was less than 1%, and the volume change needed for full bending dropped to about one-tenth of the conventional design. The actuator also moved from a straight shape to a quasi-circular shape in 50 ms at $\Delta P = 345$ kPa and operated for more than 106 cycles without substantial degradation [51]. This result is important because it shows that chamber architecture controls not only curvature, but also air demand, strain level, response speed, and durability.

The pressure-volume results from the same study are more informative than bending angle alone. Under matched material conditions, the conventional PneuNet needed about three times higher pressure, eight times greater volume change, and thirty-five times more input energy than the fast PneuNet to reach full bending. It also dissipated nearly thirty times more energy per cycle [51]. These numbers show the real difference between the two designs. Two actuators may reach a similar bending shape, but the energy cost can be very different. The study also showed that bending can change at high pressurization rates, which means that valves, tubing, venting, and supply dynamics can affect the measured response even when the actuator geometry is the same[51].

Later studies extended the PneuNet concept by modifying chamber orientation and spatial layout to create more complex motion. Oblique chambers added directional bias, allowing bending and twisting to occur together without a separate torsional actuator [52]. Three-dimensional PneuNet actuators used spatial chamber arrangements and tailored stiffness distributions to produce more complex movement [53]. Multigait soft robots showed another use of the same principle, where distributed chambers generate different locomotion patterns through pressure sequencing [54]. These studies show the flexibility of PneuNets, but they also reveal an important trade-off. As the chamber layout becomes more complex, the actuator becomes more sensitive to wall thickness, local stiffness, sealing quality, and fabrication accuracy. Small changes in these factors can shift the strain field and change the final motion.

PneuNets show that bending in soft pneumatic actuators can be programmed effectively through chamber architecture and compliance distribution, but they also demonstrate that curvature alone is an incomplete basis for evaluation. Their principal strength lies in the relative simplicity with which patterned chambers can generate programmed bending under a single pressure input. Their principal limitation lies in the fact that useful motion, pressure demand, volume change, hysteresis, and durability are tightly coupled to geometry and supply conditions.

### 3.2 Fiber-reinforced bending soft actuators

Fiber-reinforced bending actuators use inextensible fibers to control how a pressurized elastomeric chamber deforms. Without fiber reinforcement, the actuator body tends to expand radially, behaving more like a simple balloon than a source of useful directed motion. Fibers restrict this expansion and redirect the deformation into a more controlled bending response [55], [56]. The important point is that bending is not produced by pressure alone. It depends on how the reinforcement layout redistributes strain within the actuator wall.

Two design strategies are common in this actuator family. The first creates structural asymmetry, usually through a strain-limiting layer or a bilayer arrangement, so one side elongates more than the other during pressurization [56]. The second uses multiple chambers, where differential pressure generates the bending moment. Both strategies aim to convert radial inflation into useful curvature while limiting uncontrolled expansion. This separates fiber-reinforced bending actuators from PneuNets. PneuNets rely mainly on localized chamber inflation, whereas fiber-reinforced designs use external or embedded constraints to guide the deformation more mechanically [49], [56], [57].

Fiber orientation is the main design variable, but it is not the only one. Circumferential or near-circumferential fibers strongly limit radial expansion and can promote axial extension. As the fiber angle changes, radial expansion becomes less restricted, and torsion may appear because of the helical reinforcement path [57], [58]. This creates an important design problem: one fiber family can produce both useful and unwanted deformation. A bending actuator may twist, and an actuator intended for extension may develop curvature. Symmetric fiber layouts are therefore often used to reduce parasitic torsion while preserving the required bending or axial response [58]. In this sense, reinforcement is not only a strengthening feature; it is part of the actuator's motion design.

Model-based design shifted this actuator family from trial-and-error geometry selection toward motion-driven design. Connolly and co-workers treated the desired actuator motion as the starting point and used analytical modeling with optimization to select fiber angle, segment length, and related geometric parameters [41]. This is valuable for wearable systems, where the actuator must follow joint-like motion rather than simply bend as much as possible. Still, the method has limits. Its accuracy depends on assumptions that are difficult to maintain in fabrication, such as uniform wall thickness, accurate segment length, and consistent material behavior [56], [57], [58]. Optimization can reduce trial and error, but it cannot remove manufacturing variability.

That variability remains a major weakness of fiber-reinforced bending actuators. The actual response depends on fiber orientation, winding tension, wall-thickness uniformity, bonding quality, and the fiber–matrix interface. Small errors in these details can change the pressure-curvature response, introduce unwanted deformation, and reduce repeatability. This is why fiber angle should not be treated as a complete design solution. It is a useful control parameter, but predictable performance also requires accurate fabrication and stable bonding during repeated actuation [44].

Wearable-oriented designs show both the value and the limits of this approach. A three-air-chamber actuator reinforced with a double-helical Kevlar layer can bend in different directions by controlling pressure in the chambers (Figure 2c) [55], [59]. The reinforcement helps limit radial expansion while keeping the chamber structure relatively simple. However, the main question is not only whether the actuator bends. It is whether it bends repeatably under realistic load, with acceptable pneumatic demand and controllable error. In the reported wearable design, bending-angle errors remained noticeable, fingertip forces were about 0.5 N, and the authors recommended limiting the operating pressure to reduce modeling error at higher pressures [59]. This is an important lesson: increasing pressure can increase motion, but it can also amplify nonlinear behavior, constraint imperfections, and material losses.

Manufacturing strategy is the main source of the viability of this actuator family. In several recent studies, reinforcement is embedded between silicone layers through repeated casting and wrapping steps to improve placement stability and reduce fiber migration during use [59]. This can improve structural consistency, but it also increases process complexity and introduces additional sources of variability, including layer alignment, encapsulation quality, and interface defects.

Fiber-reinforced bending soft actuators show that bending can be programmed effectively through reinforcement-mediated anisotropy, but they also illustrate the limits of relying on nominal geometry alone as a predictor of performance. Their main advantage lies in the ability to direct strain mechanically and to suppress undesired radial deformation more effectively than many unconstrained designs. Their main limitation lies in the fact that bending behavior remains strongly dependent on reinforcement architecture, manufacturing precision, material nonlinearity, and loading condition.

### 3.3 Fabric-based pneumatic bending actuators

Fabric-based pneumatic bending actuators generate curvature by inflating a thin bladder inside a textile structure. The bladder supplies the pressure-driven expansion, while the fabric layers, seams, folds, and elastic elements decide where that expansion can go [60], [61]. This is different from silicone PneuNets, where the chamber geometry and local wall compliance are built into a molded elastomer. In fabric actuators, much of the constraint is moved into the textile assembly itself. Without this constraint, the bladder would simply balloon outward; with it, inflation can be redirected into bending, straightening, or more complex deformation.

Early studies on fabric-based actuators were important because they offered a practical way to build low-profile wearable robotic systems; it was practical. Cutting, sewing, laminating, and bonding flat textile layers made it easier to build low-profile actuators for gloves and wearable devices [60]. A latex or flexible bladder could be placed between fabric layers, and the actuator shape could be changed by modifying the textile pattern instead of redesigning a mold [60], [61]. For wearable systems, this matters. Comfort, thickness, fit, and ease of customization can be as important as bending angle or output force. The weakness is also clear: textile behavior is variable. Weave or knit direction, seam compliance, sealing quality, and bonding all affect the pressure–curvature response. This makes fabric actuators easier to adapt, but harder to reproduce consistently than many monolithic silicone actuators.

A key step in this field was the intentional use of textile anisotropy. Cappello et al. showed that knit, woven, and nonwoven layers could be combined and oriented to produce bending, straightening, and multi-degree-of-freedom behavior at relatively low pressure while keeping a thin wearable form [62]. This changed the role of fabric from a passive covering to an active constraint layer. The actuator response could be adjusted through fabric type, fiber direction, seam layout, and layer sequence [63]. This gives fabric actuators a wide design space, but it also makes the design harder to describe with a few simple parameters. Two actuators with the same nominal pattern may behave differently if the textile stiffness, seam quality, or bonding changes. For this reason, textile characterization and process control are not secondary details; they are part of the actuator design [12]

Wearable glove studies show the real strength of this actuator class. In rehabilitation and assistive gloves, the main goal is not maximum curvature. The actuator must help grasping while remaining light, safe, and comfortable during use[64]. Fabric actuators can lie closer to the body, tolerate some alignment mismatch, and fit more naturally into garments than many bulkier silicone actuators. However, performance claims in these systems need careful reading. Some demonstrations still rely on off-board pneumatic supplies, simplified control, or test conditions that do not fully represent daily wearable use [64], [65]. In my view, this is one of the main comparison problems in wearable soft actuation: the reported function often reflects the actuator, pneumatic supply, tubing, and control strategy together, not the actuator body alone.

Recent studies have moved in two useful directions. The first is parameter-based and model-assisted design. Ge et al. related textile selection and geometry to blocked force and torque, then used different actuator types in a wearable glove for hand motions [66]. Nguyen and Zhang used finite-element analysis to predict bending and tip force in several fabric actuator classes and discussed bandwidth and efficiency trends [63]. These studies are important because they move fabric actuator design beyond trial and error. Still, the models remain sensitive to textile nonlinearity, seam mechanics, bonding compliance, and pneumatic boundary conditions. A model may work well for one textile and fabrication route, but that does not mean it will transfer easily to another.

The second direction is stronger constraint programming through folds, pleats, and elastic elements. Pleated textile actuators use fold geometry to concentrate deformation and can achieve full bending at lower pressure than comparable fiber-reinforced silicone devices, while also producing higher blocked force at the same operating point [67]. Suulker et al. showed that adding a pre-tensioned elastic band could improve bending and output force in a hand-exoskeleton actuator [68]. These changes improve performance, but they also introduce new failure concerns. Pleats, stitches, and elastic inserts can introduce hysteresis, local stress concentration, stitch fatigue, and wear-sensitive behavior [67], [68]. Device-level systems such as Exo-hand show the benefit of lighter and more compact textile actuation, but they also show that comfort, fit, donning, and daily usability can decide the success of the device as much as actuator-level force or angle [69].

The newest work is moving toward programmable and customizable textile actuation. Tanaka et al. used patterned reinforcements, including Turing-like textures, to modify inflation pathways by redistributing constraints over the textile surface[67]. Guo et al. developed encoded sewing, where seam placement and heterogeneous stretch guide three-dimensional inflation from a flat sewn shell [70]. This is a promising direction because the fabrication process itself becomes part of the actuation design. At the same time, it raises the standard for manufacturing control. Sewing precision, seam stiffness, textile consistency, sealing quality, and batch-to-batch repeatability become central if these actuators are to move beyond one-off prototypes.

Fabric-based pneumatic bending actuators show that bending can be programmed effectively through textile-mediated constraint architectures that combine low profile, low mass, and scalable two-dimensional fabrication. Their main advantage lies in the systems-level benefits they offer for wearable applications, including comfort, packability, and ease of customization, while recent designs increasingly achieve useful bending at lower pressure and with improved packaging efficiency [60]. Their persistent limitations lie in variability and durability mechanisms associated with textiles and seams, including anisotropy drift, stitch fatigue, delamination, leakage, and dependence on fabrication quality [12].

### 3.4 Bending soft sleeve actuators (BSSAs)

Bending soft sleeve actuators (BSSAs) are a class of pneumatic bending actuators that are implemented in a sleeve architecture with internal constraints that intentionally create an asymmetry in the pneumatic expansion that produces the desired bending [23]. Rather than relying on unconstrained wall bulging to produce deformation, BSSAs use a folded bellows structure in combination with tie restring -layers that locally inhibit expansion. The most developed designs involve a dual-chamber sleeve that shares a compliant folded actuator frame, with the position of constraining layers dictating which side of the actuator can be extended more freely. Therefore, the pressure will cause asymmetric deformation and a bending moment towards the unconstrained side, and by switching the pressurization of the chambers, the bending moment will be bidirectional [23].

In the early BSSA designs, a deformable corrugated outer wall and a stiffer flat inner wall were used to provide unequal axial extension through the actuator thickness [23]. However, this design showed a major limitation: the rigidity of the inner wall reduced actuator flexibility and required higher actuation pressure. Instead, a better way was to achieve the desired asymmetry by introducing local constraining layers rather than relying on a rigid inner wall. In this design method, non-flexible constraints are imposed only where expansion needs to be restricted. This enables the actuator to retain more compliance while providing directionally bending [23].The lateral-support model resulted in better bending moment generation but could only be used for bending angles of less than 40°. The BSSA with the longitudinal support showed better performance. With the combination of many folded triangular walls, longitudinal tie-restraining layers and end-placed constraining layers, it has reached a bidirectional bending of up to 70° per side or 140° in total [23].

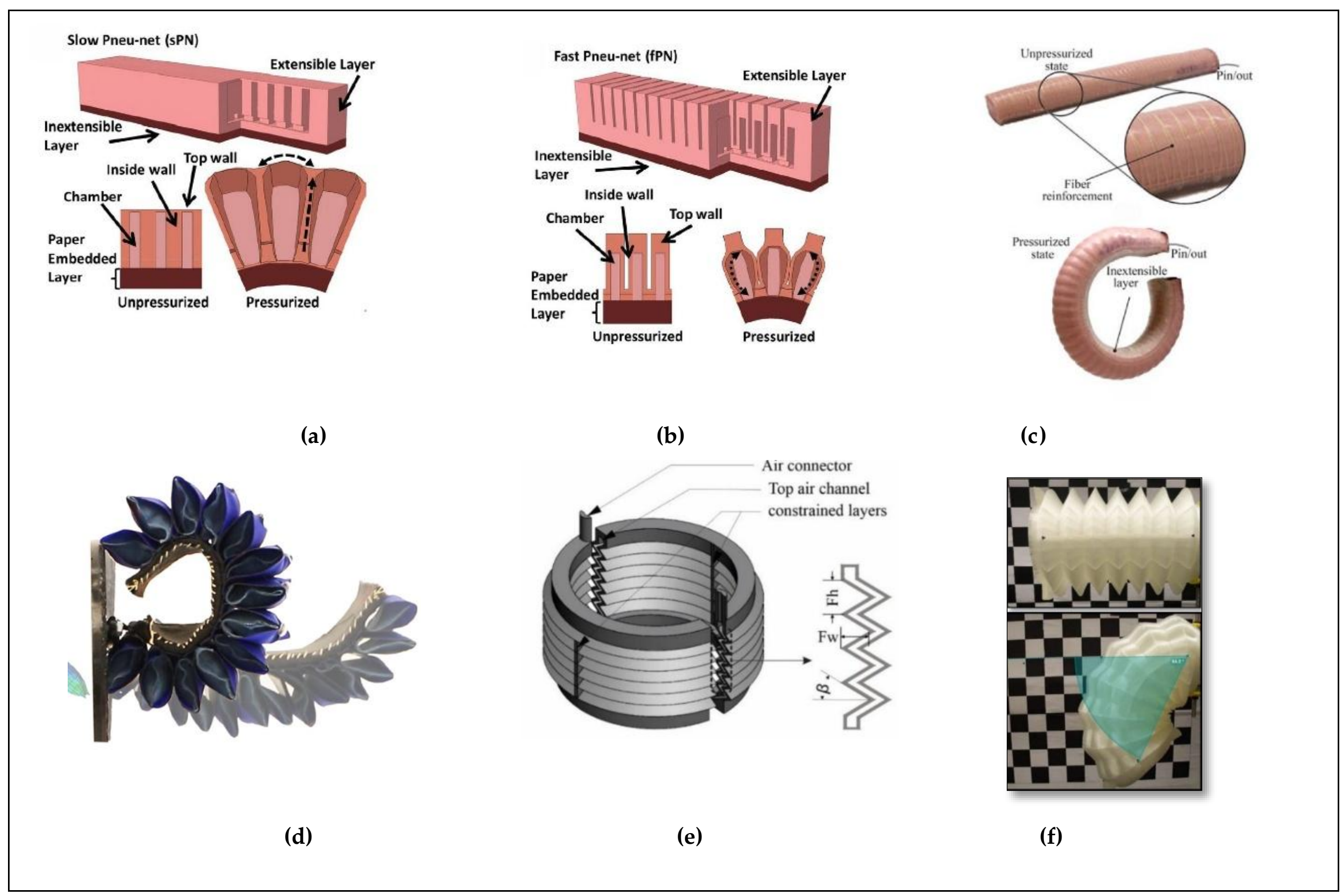

**Figure2:** Bending actuator: (a) Pneu-net [51], (b) Fast Pneu-net [51], (c) Fiber-reinforced [56], (d) Fabric actuator[63], (e) BSSA concept [23], (f) BSSA prototype [23].

At the system level, BSSAs offer practical advantages compared with many bending actuator classes. They are not only bending elements, but they are sleeve structures that can be placed around a body segment directly [23], [24]. This is important in wearable devices, as external straps, braces, and textile attachment layers can stretch, dissipate force and decrease comfort with repeated use. BSSAs can minimize the interface losses and provide a more compact approach to providing bending assistance, by placing the load path within the actuator body. Dual chamber also offers a bidirectional bending capability within a single integrated actuator, rather than multiple antagonistic actuators or extra mounting parts.

Although BSSAs address some wearability limitations of conventional bending actuators, they also have a major limitation. First, BSSAs have a high sensitivity to the constraints within them. A slight modification to the position, stiffness, or symmetry of the constraining layer can alter the direction of bending, decrease repeatability, or cause the actuator to be too stiff. The key design challenge for BSSAs is finding the right level of constraint to prevent uncontrolled deformation, while simultaneously retaining the compliance required for wearability. Furthermore, the fabricating of thin, air-tight, folded thermoplastic sleeves, is also difficult to replicate consistently.

## 4. TWISTING ACTUATORS

Soft pneumatic twisting actuators generate axial rotation by converting pressure-driven expansion or vacuum-driven collapse into distributed shear along the actuator body. Because an axisymmetric soft chamber under pressure tends to bulge and elongate rather than twist, torsion requires a deliberate source of anisotropy or chirality. Twisting is therefore not a natural consequence of pressurization alone, but a programmed response created by reinforcement, chamber geometry, folds or cuts, pre-strain, or internal constraint architectures [71]. From the perspective of this review, twisting actuators are best compared according to the strategy used to bias deformation toward helical rotation while suppressing competing modes such as bending, axial extension, or uncontrolled radial expansion.

### 4.1 Reinforcement-programmed torsion

Reinforcement-programmed torsion is an actuation approach in which the actuator is not allowed to inflate freely. It is formed by wrapping a helical fiber, braid or cloth sleeve around the elastomeric chamber to limit the chamber's circumferential expansion and to force some of the wall strain into axial rotation (Figure 3a). The fiber angle is the main design variable. Changing this angle, or the braid pitch, can shift the actuator response from mostly axial extension to twisting motion [58], [72]. The main idea in this actuator family is that torsion is produced by the constraint pattern, not by chamber inflation alone.

The benefit is that it is relatively clean twisting motion that ballooning is less pronounced compared to many chamber-only designs. However, this class is sensitive to fabrication quality. Twist angle and torque depend on fiber angle, braid tension, sleeve thickness, bonding quality, and friction between layers. Reported designs can achieve large rotation and measurable torque, including 116.7° of twist and 0.81 N·m at 300 kPa, but these values should be read together with pressure, torque, hysteresis, and repeatability [73], [74].

A large twist angle does not necessarily indicate good performance. If the reinforcement slips, stretches, or separates from the elastomer, the pressure-twist response changes and cyclic reliability decreases. For this reason, reinforcement-programmed torsion is most effective when the constraint layer is produced accurately and remains stable under repeated loading.

### 4.2 Chamber-geometry-programmed torsion

Another method of producing twisting is to design the chamber geometry so as to create a torsional bias. These actuators have oblique or helical chambers, and cause unequal axial strain around the actuator body, which results in a twisting moment about the axis when inflated (Figure 3d). For oblique PneuNet-type designs, the chamber angle is the main design variable. Wang et al. showed that the same actuator structure can produce twisting about one axis and bending about another, depending on chamber orientation and loading direction [52]. This is useful, but it also shows the main weakness of this approach: torsion can easily appear with bending unless the chamber layout is carefully balanced.

The concept is used more directly in helical-chamber actuators, which include spiral channels around the body of the actuator. This fits well with molding and additive manufacturing because the twisting response is built into the printed or molded geometry rather than added through an external sleeve [75]. Vacuum-driven designs are similar, but they rely on collapsing inwards along a helical path to minimize some of the outward bulging of positive-pressure actuators. In my view, the main strength of this family is integration. The actuator body and the torsional mechanism are made as one structure. The limitation is motion coupling. Without an external reinforcement layer, these actuators can still show radial expansion and coupled bending or extension [74].

### 4.3 Thin-sheet and laminated architectures

Thin-sheet and laminated twisting actuators take a different path. They do not rely on a thick chamber or an external reinforcement but rely on a large number of miniature inflation units glued into a thin laminate (Figure 3c) [76], [77].These units expand together causing a net twist. Gorissen et al. showed this clearly with a pneumatic balloon actuator array, which reached about 173° of rotation in 6.5 s at 130 kPa [77]. This result is useful because it shows how large angular motion can be obtained from a light and flat structure, not from a bulky torsional body.

The weakness is also obvious. These actuators are high-performance and can be highly rotatable, but generally they do not have much torque or durability. These actuators depend on thin membranes and bonded layers that are repeatedly inflated. Because of this, delamination, adhesive creep, and low torsional stiffness may limit performance more than the twist angle itself . Fabrication is also difficult to control because many small bonded elements must deform together. For this reason, these actuators are often described using experimental pressure–angle curves rather than broad design rules [77], [78]. I see this class as useful mainly when the actuator must be light, thin, and easy to fabricate. However, it is less suitable when the application needs high torque, long cycle life, or strong torsional loading.

### 4.4 Fold-guided torsion

Fold-guided torsion is an approach where the axial deformation is linked to the rotation through the fold pattern [71]. The Kresling pattern is the most obvious one since it operates as a bellows which twists as it expands and contracts (Figure 3b). This is appealing, as evidenced by the designs reported, where one Kresling-based actuator was able to achieve 250° of rotation under 50 kPa of vacuum and another origami-structured actuator generated approximately 435° over a relatively low pressure range [79]. These results show that large rotation can be achieved without relying mainly on bulk shear deformation.

Main limitation is durability. Folds are associated with high local strain and repeated folding may lead to fatigue, leakage and hysteresis. Also, Repeated contact between the folded surfaces can alter the pressure–rotation response during cycling. The fold pattern makes the motion easier to model in principle, but accurate prediction still depends on crease stiffness, friction, and pressure–volume behavior. Fold-guided torsion is useful for producing large rotation at low pressure, but its real limitation is not the twist angle. It is whether the folds remain sealed and durable during repeated use.

### 4.5 Pre-twist programming

Pre-twist programming uses a simpler idea. The initial twist of the actuator is applied prior to pressurizing, which means that the starting shape already has a twist (Figure 3e). When pressure is applied, the tube releases or amplifies this stored twist and produces rotation. Two pre-twisted tubes can also be placed back to back to create both directions of rotation, or to fine-tune the stiffness. In one of the designs reported, the actuator could achieve up to 120° rotation at 130 kPa [80] .

the strength of this actuator is its mechanical simplicity, which does not require complicated chamber designs, multi-layer reinforcement or fold patterns. However, its main weakness is long-term stability. Since the motion depends on stored pre-strain, material creep and viscoelastic relaxation can shift the pressure–angle response during repeated use. Hysteresis and rate effects are also difficult to avoid.

### 4.6 Sleeve-based torsional mechanisms

The difference is that sleeve-based torsional actuators are not mainly designed to achieve the largest twist angle, but to make twisting practical in compact wearable systems. In these actuators, folds and internal supports help the sleeve twist as intended while reducing radial expansion and pneumatic volume (Figure 3f).

The twisting soft sleeve actuator (TSSA) is a representative example [24]. It uses helically folded bellows to define the twisting path, while an internal stabilization layer connects the inner and outer walls and limits outward bulging. This allows the sleeve to deform in a more coordinated way. Under positive pressure, the actuator twists in one direction; under vacuum, it twists in the opposite direction. The reported performance was about 30° of clockwise rotation and a peak output force of 40 N at 75 kPa [24].

The main value of the TSSA is integration. It reduces the need for bulky fixtures, supports bidirectional motion, and keeps the load path within the sleeve body. This makes it more suitable for wearable

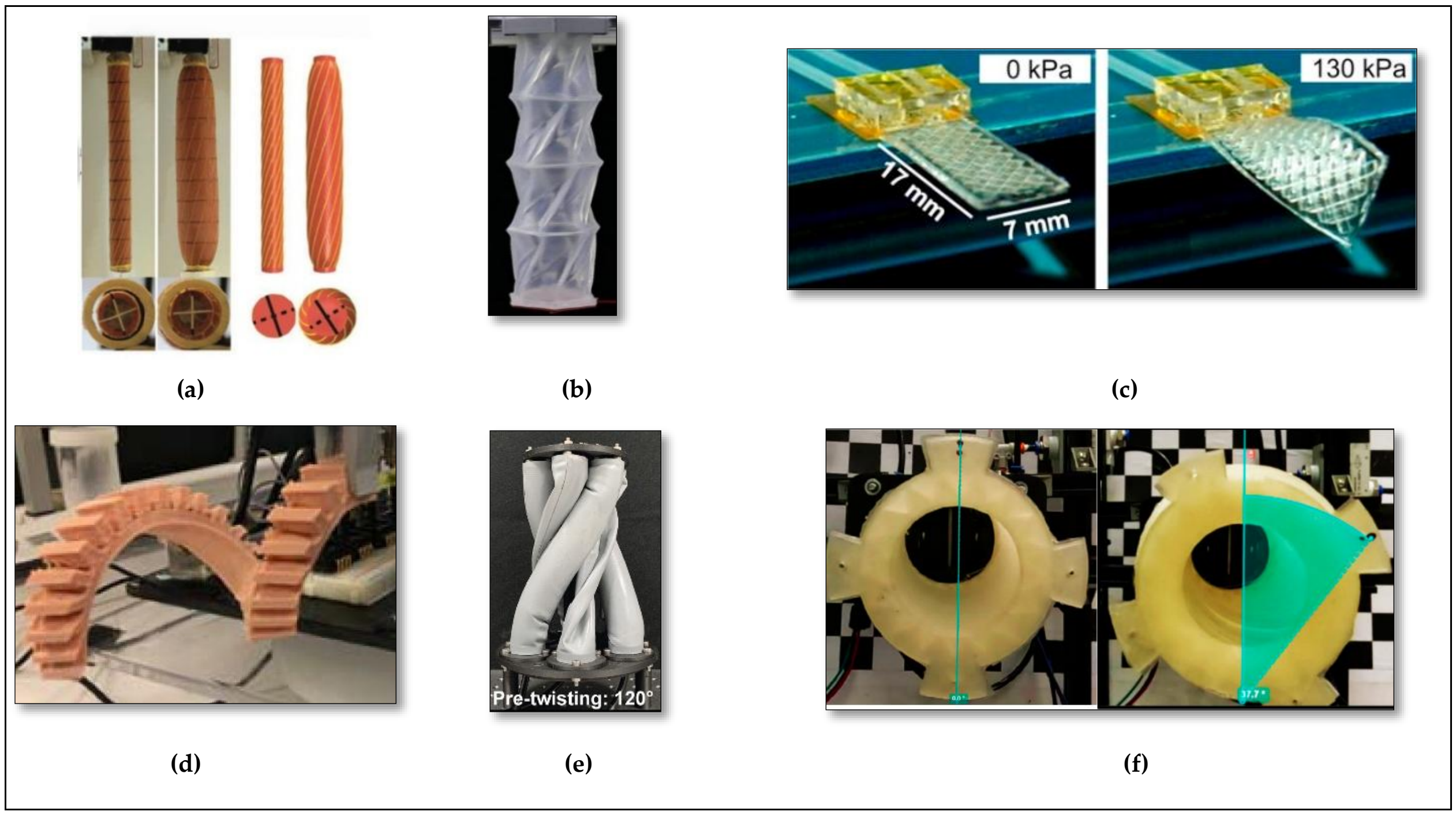

**Figure3:** Twisting actuator: (a) Fiber-reinforced [58], (b) Fold-guided [79], (c) Thin-sheet architecture [77], (d) Chamber-geometry-programmed [81] (e) Pre-twist programming [80], (f) Sleeve-based torsional mechanisms [24].

applications than many earlier twisting actuators. Its limitation is also important. The twist response depends strongly on fold geometry, internal constraint placement, sealing quality, and manufacturing accuracy. Small changes in these features can affect rotation and repeatability. For this reason, sleeve-based torsional actuators should not be judged by angular stroke alone.

## 5. OMNIDIRECTIONAL ACTUATORS

Omnidirectional soft pneumatic actuators are designed to generate a controllable curvature vector, typically by distributing multiple pneumatic chambers around the actuator axis and using differential pressurization to produce pitch, yaw, and, in many cases, axial extension [82], [83]. In the canonical three-chamber architecture, equal pressurization yields near-axial extension, whereas selective pressurization of one chamber, or one chamber pair, produces bending in a direction determined by chamber layout and the resulting differential axial strain field [84] . Across the literature, omnidirectional actuators are commonly divided into two broad classes: parallel configurations, in which multiple linear actuators are arranged externally around a central axis, and monolithic single-structure configurations, in which the chambers are integrated into a single continuous body [84], [85]. This distinction is analytically useful because it identifies where the main design burden lies. Parallel architectures trade compactness for modularity, load sharing, and straightforward actuator replacement, whereas monolithic architectures reduce assembly but become more sensitive to ballooning, inter-chamber coupling, and geometry-dependent stiffness.

### 5.1 Parallel configurations

Parallel omnidirectional actuators use several linear pneumatic elements arranged around a central axis [86], [87], often at 120° intervals (Figure 4 a, b) . If one element extends more than the others, the actuator bends in that direction. If all elements are pressurized together, the structure extends axially. The main appeal is modularity. Force, stiffness, and repairability can be adjusted by changing the number, type, or arrangement of the elements.

This principle was established early in continuum robots, where three longitudinal pneumatic elements were arranged around a centerline and controlled through differential pressure [88]. Air-Octor showed the large workspace possible with this layout, but it also showed the cost of this design: several pressure lines, external supports, and fabrication differences can create coupling and calibration drift [88]. Later systems, such as OctArm, improved the response by using segment-level actuation and external constraints, including meshes or skeletons, to reduce radial expansion and make bending more predictable[89]. STIFF-FLOP added another layer by combining pneumatic actuation with tendons or granular jamming to improve stiffness and stability under load [90]. In medical and inspection robots, the same parallel layout is useful because it can keep a central lumen while supporting controlled bending in confined spaces [91].

In my view, the strength of parallel configurations is not compactness. It is modularity, redundancy, and load sharing. Their weakness is system complexity. More chambers usually mean more tubes, valves, regulators, and alignment issues. These factors increase size and make performance harder to compare because bandwidth, force, and repeatability may depend on the pneumatic hardware as much as the actuator geometry. For this reason, parallel configurations are most useful when load capacity and modularity justify the added complexity. For smaller integrated systems, monolithic or quasi-parallel bodies are often more practical.

### 5.2 Monolithic single-structure configurations

Monolithic omnidirectional actuators are made from a single continuous body with radially arranged chambers (Figure 4c).The biggest benefit is integration: they minimise the number of parts, the complexity of packages and assembly time, since they are not bundled externally. Most common issue is ballooning. During radial expansion, the air provided is used for swelling and not the bending, thus decreasing the pneumatic efficiency and the pressure–curvature response (Figure 4d).

The typical design consists of a cylindrical or annular body containing three or more longitudinal chambers which encircle a central lumen [82], [92]. Differential pressurization bends the actuator, while equal pressurization can produce axial extension. This layout is simple, but it is not necesserly reliable. The pressure–shape relationship may be affected by variations in wall thickness, end constraints, material hysteresis and shared chamber walls. To this end, recent research has shifted toward parameterized design, by which the variables of chamber angle, wall thickness, lumen size, and others, are correlated with the curvature, force, and workspace by simulation and optimization [93]. In my view, this is an important shift because the problem is no longer only how to create multidirectional bending, but how to make that bending predictable.

Bellows and corrugated geometries offer another solution. Instead of allowing the wall to swell uniformly, folded walls make the actuator deform through fold opening, which can reduce ballooning [83] [84]. This is a useful mechanism-level improvement because it controls deformation through geometry rather than external reinforcement. The trade-off is durability. Fold roots and end transitions can concentrate strain, so fatigue, cracking, and leakage become important design limits [84]. These actuators should therefore be judged by workspace and curvature, but also by cyclic durability and the stability of their pressure–shape response.

Control-oriented designs make this issue even clearer. The 3D-PSA, for example, combined geometric optimization with pressure-to-shape mapping and static and dynamic testing [82]. This type of work is valuable because it treats omnidirectional actuation as a mapping problem, not only a geometry problem. The best design is not always the one with the largest workspace. It is often the one that gives a repeatable and low-variation relationship between pressure input and shape output.

Some monolithic designs add backbones, inextensible hoses, or sealed chamber units to stabilize the neutral axis, improve sealing, and reduce unwanted torsion [84]. These features can improve repeatability, especially in longer continuum manipulators, but they also increase stiffness and diameter. Other designs use anisotropic cross-sections, such as triangular layouts, to reduce ballooning without separate reinforcement. This can simplify fabrication, but it may also introduce direction-dependent stiffness and nonuniform bending [84].

Overall, monolithic omnidirectional actuators offer a compact route to multidirectional motion, but they are sensitive to ballooning, chamber coupling, and geometry-dependent control errors. Their main design challenge is not simply to bend in many directions. It is to produce a stable pressure–shape response while limiting radial growth and maintaining durability.

### 5.3 Sleeve-based omnidirectional actuators

Sleeve-based omnidirectional actuators move the monolithic concept toward wearable use. The omnidirectional soft sleeve actuator (OSSA) is a representative example [94]. It uses a folded bellows sleeve divided into four sealed chambers by internal sealing layers and longitudinal tie-restraining elements (Figure 4f). These internal constraints keep the inner and outer walls aligned during actuation. Differential chamber pressurization produces bending in different directions, common pressurization produces axial extension, and vacuum produces active contraction. In this design, ballooning is not controlled by an external frame. It is reduced through the sleeve structure itself, mainly by the folded wall geometry and internal constraint coupling.

The value of the OSSA is integration. It combines multidirectional bending, extension, and contraction in one compact sleeve. This helps address problems seen in parallel and conventional monolithic actuators, such as large diameter, external stabilization, and wasted

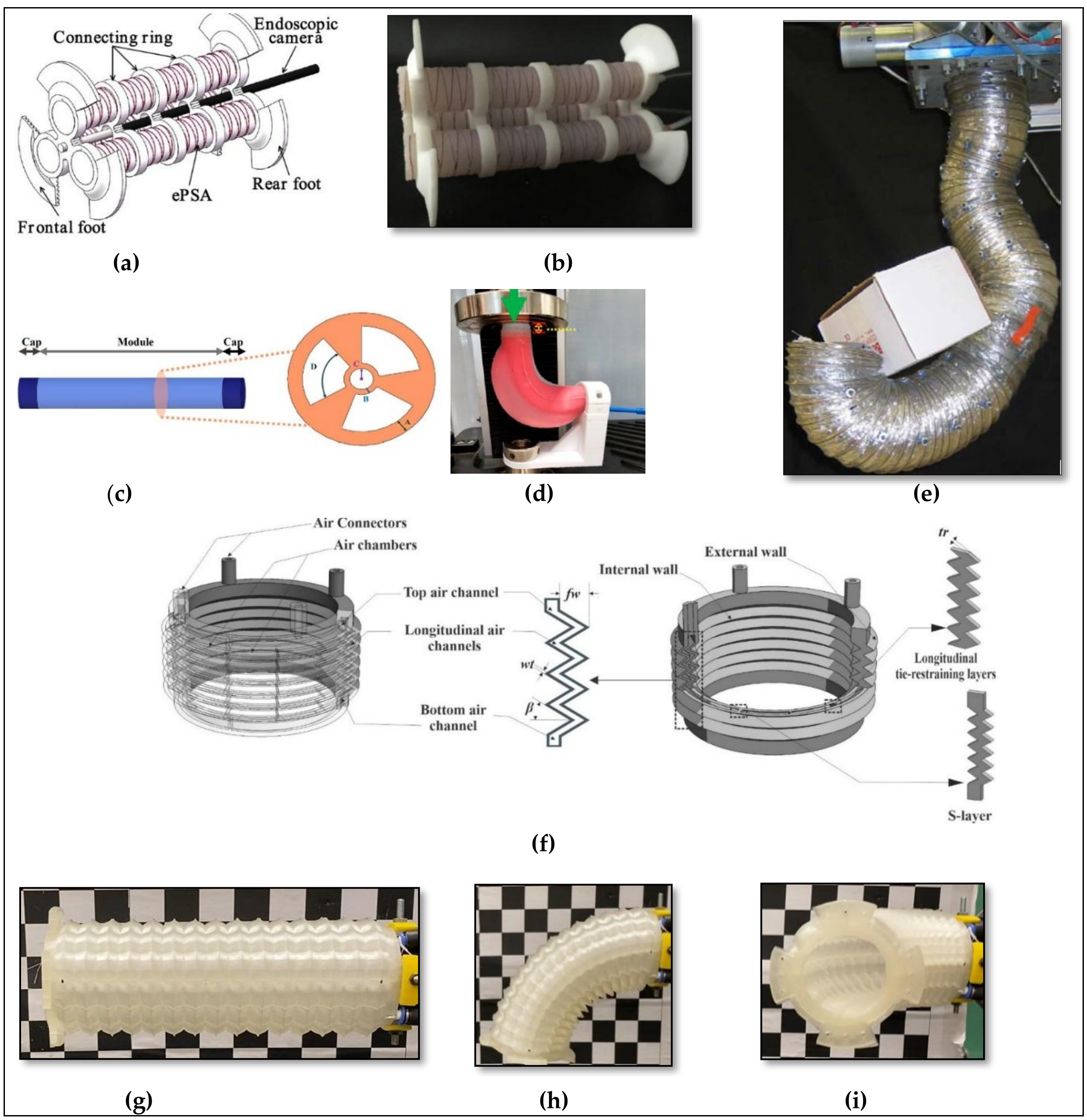

**Figure4:** Omnidirectional actuator: (a) Design of parallel configuration actuator[95], (b) parallel configuration actuator [95], (c) design Single structure actuator [96], (d) Single structure actuator [96], (e) Air-Octor concept [89], (f) design of OSSA[94], (g) OSSA extension motion [94], (h, i) OSSA bending motion[94] radial expansion. The limitation is manufacturing sensitivity. Sealing quality, fold-root durability, constraint symmetry, and boundary conditions can all change the actuator response. This point is especially important for wearable use, where straps, contact with the body, and fixation method may affect the pressure–shape relationship. Sleeve-based omnidirectional actuators should not be judged only by workspace or bending angle.

## 6. DISCUSSION

this review shows that soft pneumatic actuator design should be judged by how effectively the structure converts pressure into useful deformation while limiting unwanted deformation, pneumatic demand, and system-level penalties. The studies reviewed indicate that the nature of the soft pneumatic actuator performance depends primarily on the constraints that are applied within each architecture. Grouping actuators by motion type is useful, but it does not explain performance by itself. PAMs, bellows, FREEs, and LSSAs can all produce axial motion, but they rely on different constraints: braided sleeves, pleated membranes, corrugated walls, fiber orientation, or tie-restrained folded sleeves. The same applies to bending actuators. The localized chamber restraining is applied to PneuNets, anisotropic reinforcement is applied to fiber-reinforced actuators, fabric constraints are applied to fabric actuators, and internal constraining layers are applied to a sleeve body in BSSAs. These mechanisms can generate similar movement, but they are vastly different in terms of the pressure they require, the power they generate, their repeatability, durability, and how they are fabricated.

Several studies show that adding constraint improves motion directionality, but it often reduces compliance or increases design sensitivity. Although Braided PAMs provide high tensile force, bladder–sleeve friction, braid geometry, and end effects all affect its response. Pleated PAMs addressed some of these issues, yet they shift the durability concern toward fold quality. Bellows show positive outcomes in reducing radial ballooning and enhancing axial motion but, but their performance becomes limited by off-axis deformation, buckling, and local strain at the convolution roots. Fiber-reinforced actuators can guide strain more precisely, but they depend strongly on fiber angle, winding accuracy, bonding quality, and material behavior. Sleeve-based actuators make this trade-off especially clear. They reduce pneumatic volume and improve integration, but their performance depends heavily on tie placement, fold geometry, sealing quality, and manufacturing precision. One lesson from this review is that the best actuator design is not the one with the least constraint or the largest free deformation, but the one that provides enough constraint to produce the intended motion while preserving useful compliance.

The same problem is evident in the literature for bending actuators. PneuNets can bend through large angles by using a series of internal chambers; however, the bending angle alone does not measure the total quality of the PneuNets. The fast PneuNet study showed that a conventional PneuNet can reach a comparable bending shape, but it requires much higher pressure, larger chamber-volume change, and greater input energy to achieve that motion. This is a good example as it clearly distinguishes motion output from pneumatic cost. Fabric-based actuators show a different type of advantage. They are not always produced higher curvature or larger force, but low profile, comfort, and easier customization for wearable systems. However, textile actuators also introduce variability through seams, fabric anisotropy, bonding, and leakage. One approach to wearable integration problem is to incorporate the actuation structure in a sleeve form, which is a strategy employed by BSSAs, but which is very dependent on the presence of constraining layers within the structure.

The studies on twisting and omnidirectional reiterate the same message. Twisting actuators require anisotropy, chirality, folds, pre-twist, or internal constraints because a symmetric pressurized chamber will not naturally produce torsion. Reinforcement-programmed twisting can give relatively pure torsion, but interface mechanics and fatigue become important. Chamber-based twisting is more suited to printed or molded geometries but can add bending and/or extension in addition to twisting. Fold guided and pre-twisted designs may have a high rotation but do not have a high degree of durability, sealing, and drift which dampen long-term reliability. Omnidirectional actuators add another layer of complexity. Parallel designs offer good modularity and load sharing, but cause increased diameters and pneumatic overhead. Monolithic designs have better packaging, but are more susceptible to ballooning and inter-chamber coupling. The design evolution towards wearable integration is achieved by the sleeve-based omnidirectional actuators, such as OSSA-type actuators, but requires accurate sealing, fold durability and stable internal constraints.

## 7.FUTURE WORK

Future research should make the transition away from reporting motion and pay more attention to what should be done to produce the motion. The primary requirement for linear actuators is to increase force and stroke while decreasing pneumatic volume requirement and axial stiffness, and hysteresis. PAMs still need improvements to friction and nonlinear force behaviour, and bellows, FREEs and LSSAs have a need for increased data on stability under load, fatigue and repeatability. Further research is recommended for LSSAs to investigate the influence of tie-restraining layers, fold geometry, wall thickness and manufacturing quality on axial stiffness and force–displacement response.

Bending angle should not be the primary focus for bending actuators. A PneuNet, a fiber-reinforced actuator, a fabric actuator, and a BSSA may all reach useful curvature, but the pressure, air volume, force output, and fabrication effort can be very different. Future studies should include pressure-volume behavior, force under load, and actuation rate, as well as hysteresis and cyclic durability. Another focus on fiber-reinforced and fabric actuators is the manufacturing variability, such as fiber placement, seam stiffness, performance of the bonding, leakage, and textile anisotropy.

For twisting and rotary actuators, the literature should move beyond reporting only the generated twist angle. The studies should report torque, hysteresis, repeatability, and coupling with bending or axial motion. Also, Further research is needed for this sleeve-based twisting actuator in order to explore its potential for bidirectional rotation, integration on the body, and consistent torque production.

## 7. CONCLUSIONS

This review examined soft pneumatic actuators through the design strategies used to generate linear, bending, twisting, and omnidirectional motion. The main finding is that actuator performance is not defined by motion type alone. It depends on how each architecture guides pressure-induced deformation through chamber geometry, folds, reinforcement, textile constraints, or internal restraining layers. These design choices determine not only the motion output, but also force generation, pneumatic demand, repeatability, durability, manufacturability, and integration.

The reviewed literature also shows that peak motion metrics, such as stroke, bending angle, twisting angle, or workspace, are not sufficient for comparing actuator performance. However, future studies should report conditions necessary to achieve the claimed motion such as pressure range, loading condition, air volume, hysteresis, durability and fabrication quality. A better reporting practice would enable more reliable comparisons between actuators, and help design efficient, repeatable and practical soft pneumatic systems for wearable, biomedical and mobile robotic applications.